\documentclass[conference]{IEEEtran}
\IEEEoverridecommandlockouts
\usepackage{cite}
\usepackage{amsmath,amssymb,amsfonts}
\usepackage{algorithmic}
\usepackage{graphicx}
\usepackage{textcomp}
\usepackage{xcolor}
\usepackage{makecell}
\usepackage{multirow}

\def\BibTeX{{\rm B\kern-.05em{\sc i\kern-.025em b}\kern-.08em
    T\kern-.1667em\lower.7ex\hbox{E}\kern-.125emX}}
\begin{document}

\title{Dynamic Video Frame Interpolation with integrated Difficulty Pre-Assessment}

\author{Ban Chen$^1$ ~~~Xin Jin$^1$ ~~~Youxin Chen$^1$ ~~~Longhai Wu$^1$ ~~~Jie Chen$^1$
~~~Jayoon Koo$^2$ ~~~Cheul-hee Hahm$^2$ \\
$^1$Samsung Electronics (China) R\&D Center ~~~~~~~~~$^2$Samsung Electronics, South Korea \\
{\tt\small\{ban.chen, xin.jin, yx113.chen, longhai.wu, ada.chen, j.goo, chhahm\}@samsung.com}}

\maketitle

\begin{abstract}
    Video frame interpolation(VFI) has witnessed great progress in recent years. While existing VFI models still struggle to achieve a good trade-off between accuracy and efficiency: fast models often have inferior accuracy; accurate models typically run slowly. However, easy samples with small motion or clear texture can achieve competitive results with simple models and do not require heavy computation. In this paper, we present an integrated pipeline which combines difficulty assessment with video frame interpolation. Specifically, it firstly leverages a pre-assessment model to measure the interpolation difficulty level of input frames, and then dynamically selects an appropriate VFI model to generate interpolation results. Furthermore, a large-scale VFI difficulty assessment dataset is collected and annotated to train our pre-assessment model. Extensive experiments show that easy samples pass through fast models while difficult samples inference with heavy models, and our proposed pipeline can improve the accuracy-efficiency trade-off for VFI.
% This document is a model and instructions for \LaTeX.
% This and the IEEEtran.cls file define the components of your paper [title, text, heads, etc.]. *CRITICAL: Do Not Use Symbols, Special Characters, Footnotes, 
% or Math in Paper Title or Abstract.
\end{abstract}


\section{Introduction}
Video frame interpolation(VFI) aims to generate intermediate frames between
consecutive frames. VFI is widely applied in industrial products, including
slow-motion video generation, video editing, intelligent display devices, etc.
Despite recent advances in deep learning bring performance improvement, VFI models are
becoming more and more computationally expensive, making them infeasible for
practical applications. While accurate VFI models typically synthesize
high-quality results at the cost of efficiency, fast VFI models, on the other
hand, often suffer from unreasonable artifacts. 

This work aims to adjust the trade-off between accuracy and efficiency for VFI
in a data-driven manner. Based on our investigation, the performance gain of
heavy model mainly comes from challenging frames with fast moving objects or
complex texture. While for static or clean frames, VFI models always show
similar performance -- no matter what model size is. Motivated by this
phenomenon, we design a dynamic VFI pipeline, where different VFI models are
applied alternatively according to interpolation difficulty of input frames.

In our dynamic VFI, we design a VFI difficulty pre-assessment (VFI-DPA) model
based on AHIQ~\cite{lao2022attentions}, a SOTA full reference image quality
assessment (FR-IQA) model. FR-IQA aims to quantify and predict human perceptual
judgements of image quality by comparing restored images with their
corresponding pristine-quality images.  AHIQ adopts a hybrid architecture that
takes advantage of local texture CNN features and global semantic ViT features.
Our VFI-DPA employs several innovations to significantly reduce the computation
cost of AHIQ, while still improving its performance for interpolation quality
pre-assessment. To train and evaluate VFI-DPA model, we collect a large VFI
difficulty pre-assessment dataset with manual annotation.

Based on our VFI-DPA, we choose RIFE~\cite{huang2020rife} as fast VFI model and
VFIformer~\cite{lu2022video} as accurate VFI model. RIFE uses an efficient CNN flow
estimator and a simplified synthesis network to interpolate multiple frames with
real-time performance. VFIformer integrates Transformer to model long-range
dependency and handle large motions. Our VFI-DPA allows us to feed easy frames into RIFE for
efficiency, and pass hard frames through VFIformer for accuracy. 
% It achieves superior performance over RIFE,
% but is much slower.

In summary, our main contributions are as follows:
\begin{itemize}
\item To the best of our knowledge, we are the first to integrate difficulty
    pre-assessment with VFI. Our dynamic VFI enables flexible trade-off between
    speed and accuracy, which is crucial for industrial deployment.
\item We design a VFI-DPA model, a lightweight difficulty pre-assessment model to
    evaluate interpolation difficulty before running VFI models. It achieves
    promising performance with high efficiency.
\item We contribute a large-scale VFI difficulty assessment dataset with
    different resolutions and distortion types.
\end{itemize}

\begin{figure}
    \centering
    \includegraphics[height=2.53cm, width=8.51cm]{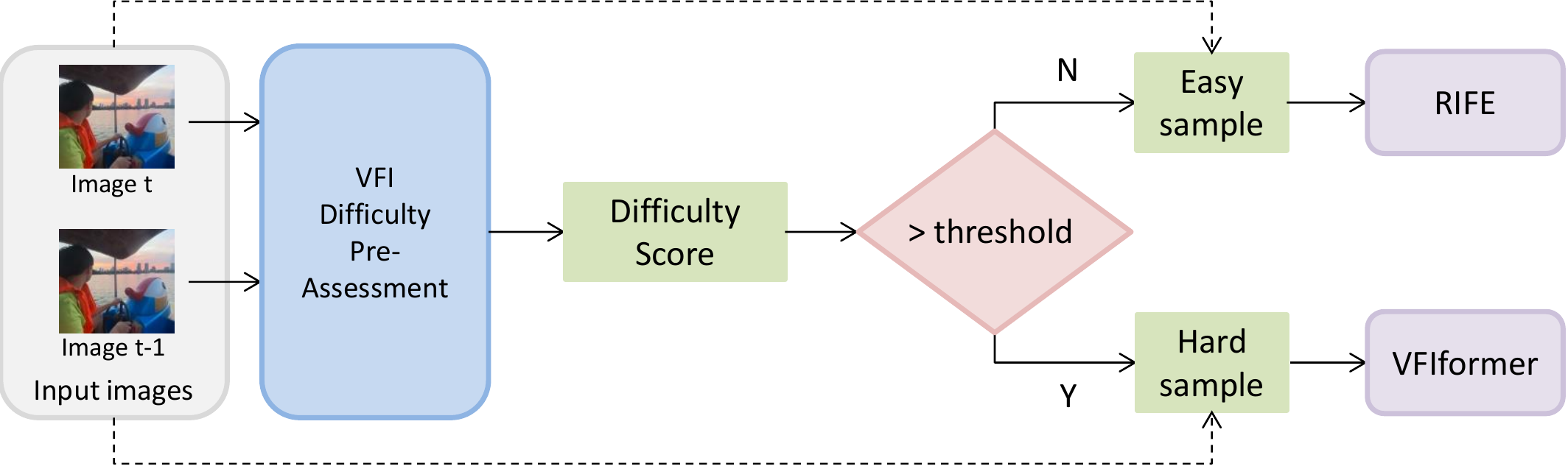}
    \caption{Overview pipeline of Dynamic VFI.}
\label{fig:dynamicVFI}
\end{figure}
\vspace{-0.01cm}

\section{Method}
In this section, we briefly introduce overall pipeline of dynamic VFI network, and then detail the architecture of VFI
pre-assessment network and its training loss. 

\subsection{Overview of dynamic VFI.} 
As shown in Fig.~\ref{fig:dynamicVFI}, we design a dynamic VFI pipeline with integrated VFI difficulty
pre-assessment.  Firstly, VFI-DPA model predicts an interpolation difficulty level for
each image pair. We then classify input frames as easy or hard class by comparing
predicted score with customized threshold. Finally, easy samples are sent to
RIFE for efficiency, while hard samples go through VFIformer for better
interpolation.

\begin{figure}[tbp]
    \centering
    \includegraphics[height=3cm, width=8.8cm]{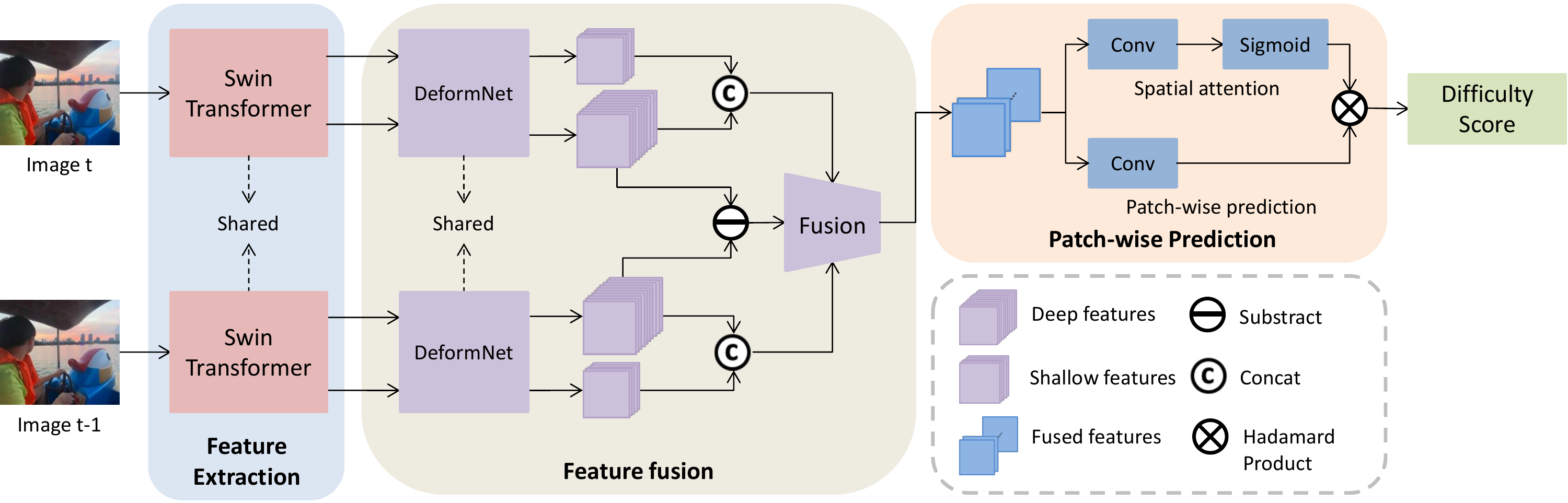}
    \caption{VFI difficulty pre-assessment model (VFI-DPA).}
    \label{fig:architecture}
\end{figure}

\subsection{VFI difficulty pre-assessment}
The overall structure of our VFI-DPA is shown in Fig.~\ref{fig:architecture}.
Given two frames, VFI-DPA predicts an interpolation difficulty score
without seeing interpolation result. It can be considered as a lightweight and
flexible plugin, which can be easily employed to choose VFI models (not
restricted to RIFE and VFIformer) dynamically based on given inputs.  VFI-DPA is
built upon AHIQ, which consists of a feature extraction module, a feature fusion
module and a patch-wise prediction module. As detailed in the following, we make
some modifications to AHIQ to make it much more lightweight but still powerful
for interpolation difficulty pre-assessment.

\begin{figure}[tbp]
    \centering
    \includegraphics[height=4cm, width=6.5cm]{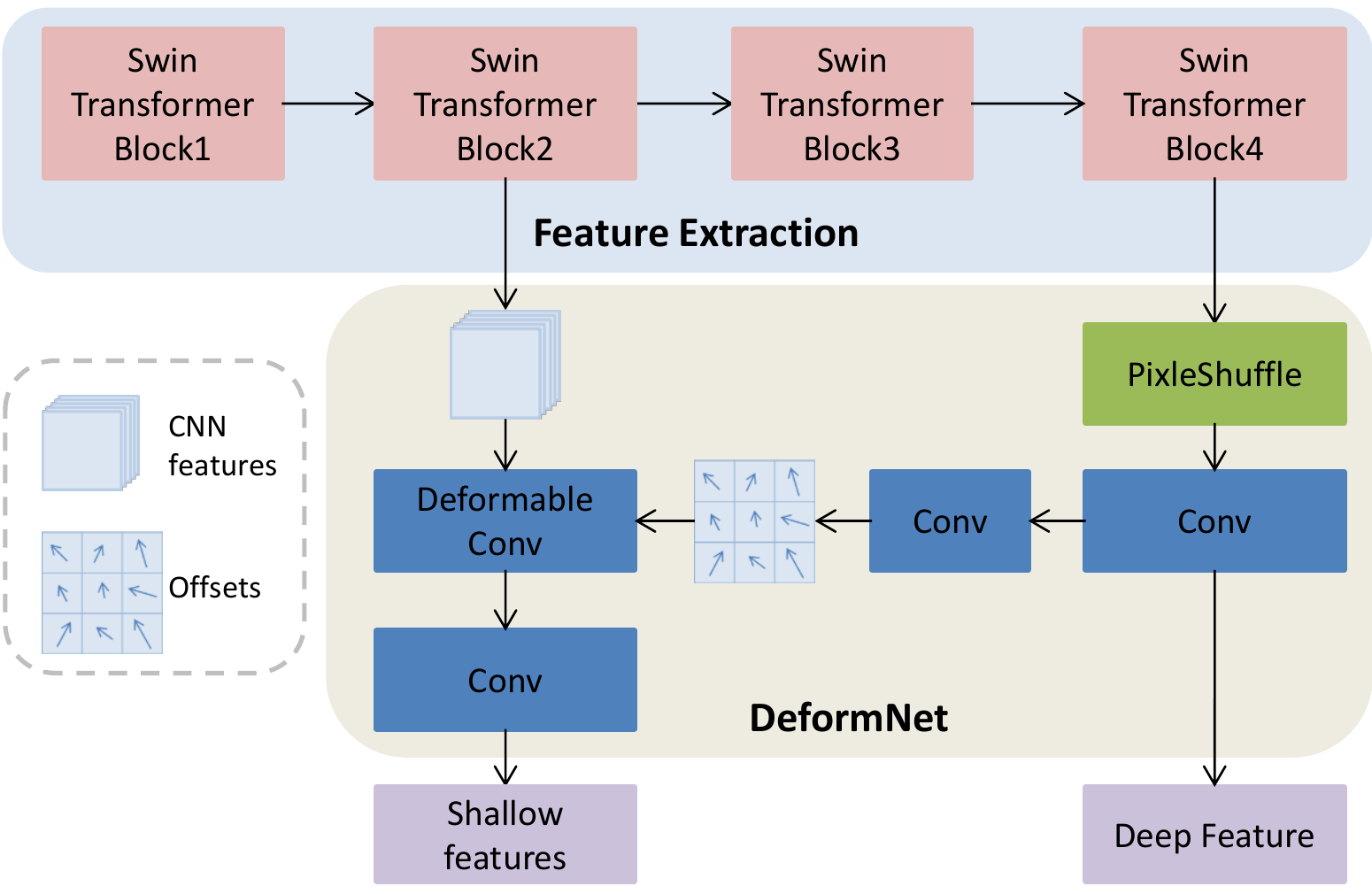}
    \caption{Feature extraction and fusion modules in VFI-DPA.}
\label{fig:fedn}
\end{figure}

\textbf{Feature extraction.} AHIQ applies a two-branch feature extraction module
that consists of a ViT branch and a CNN branch. The Transformer branch favors
long-range dependencies and global representations, while the CNN branch focuses
on local textures. While this two-branch design achieves promising results for
IQA, mixed CNN and ViT design introduces extra parameters, and is unaffordable as an
industrial-purpose backbone. On the other hand, plain ViT lacks of hierarchical
features, especially local texture features which is crucial for IQA. To solve
this issue, we use Siamese feature extraction (sharing weights) based on Swin
Transformer~\cite{liu2021swin}, where the shifted window scheme has advantage of
efficiency, and the hierarchical architecture naturally enables the feature fusion. 
As shown in Fig.~\ref{fig:architecture} and~\ref{fig:fedn},
during inference, two images are fed into feature extraction module, and then
the feature maps from Swin Transform block 2 and block 4 are outputted as
low-level texture features and high-level semantic features for further fusion.

\textbf{Feature fusion.} The feature maps from feature extraction module have
different spatial resolutions and can not be fused directly. As shown in Fig.~\ref{fig:fedn}, we first align
spatial resolution of deep feature with shallow feature by
Pixelshuffle~\cite{shi2016real}.  In contrast to coarse pooling or traditional
interpolation methods, Pixelshuffle can reconstruct high resolution feature with
less noises. Following Pixelshuffle, a 1x1 convolution layer is employed to
reduce feature channels. For shallow feature, we follow the processing in
AHIQ, by using the deformable convolution operation with offsets learned from
deep aligned feature. The deformable convolution allows the network focus on salient or moving regions in an image, which we believe is useful
to quantify the difficulty level for frame interpolation.  Then, as shown in
Fig.~\ref{fig:architecture}, we calculate temporal difference feature of
neighboring frames by subtraction and learn rich spatial representations by
concatenating deep features and shallow features within one frame. Finally, temporal and spatial features are fused by concatenation and convolution operations as final representations.

\textbf{Patch-wise prediction.} 
We follow the patch-wise prediction architecture in AHIQ. As illustrated in
Fig.~\ref{fig:architecture}, the prediction branch evaluates a difficulty score
for each location in feature map, while the spatial attention branch evaluates
the importance of each corresponding score. Finally, we calculate the final
score by weighted sum operation.

\begin{figure}[tbp]
    \centering
    \includegraphics[height=3.5cm, width=6.5cm]{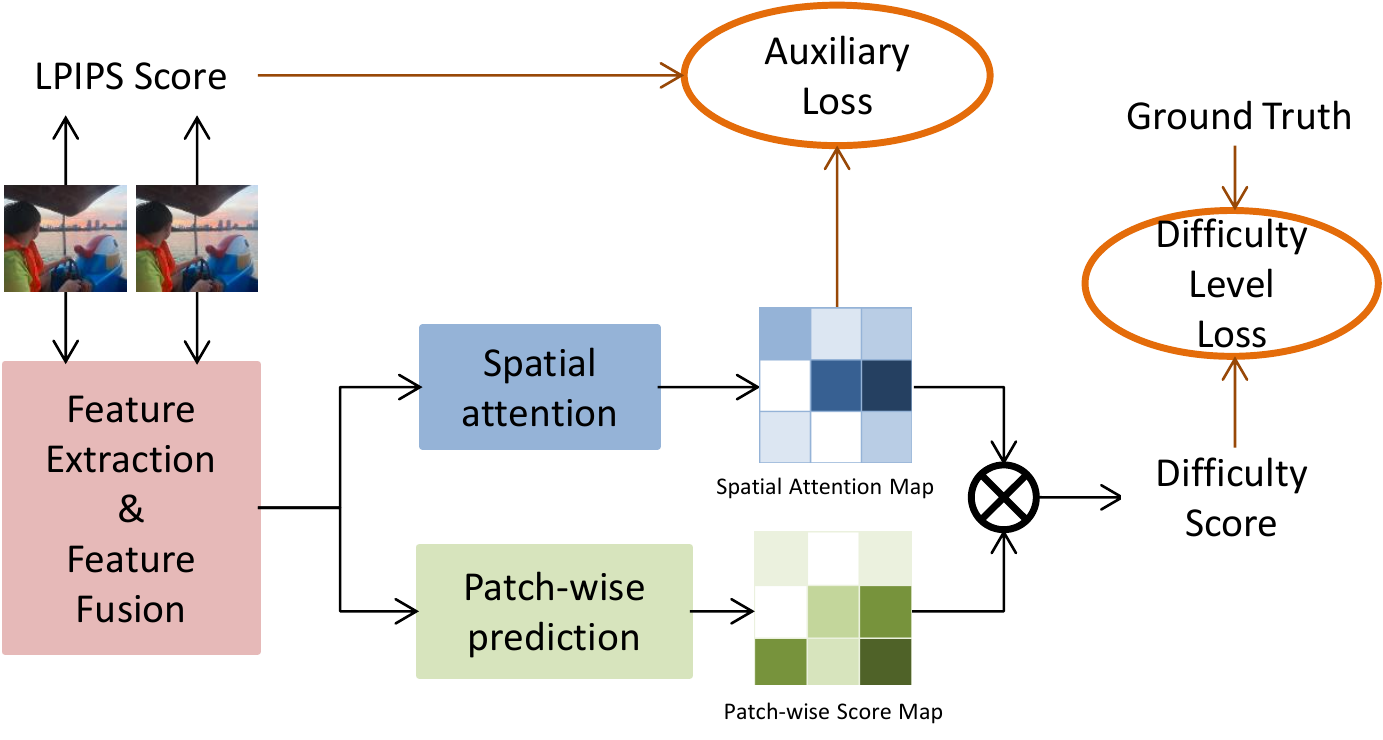}
    \caption{Detailed illustration of training loss for VFI-DPA.}
\label{fig:loss}
\end{figure}

\subsection{Training loss for VFI-DPA}
\noindent
In Fig.~\ref{fig:loss}, our loss consists of difficulty level loss between
predicted and annotated difficulty scores (normalized to $[0,1]$), and an
auxiliary loss cooperated with perceptual similarity.

\textbf{Difficulty level loss.} We adopt a difficulty level loss as
\begin{eqnarray}
    L_{difficulty} = \|  S_{GT} - S_{Pred} \Vert 
\end{eqnarray}
where $S_{GT}$ and $S_{Pred}$ are ground truth and predicted scores.

\begin{figure*}[htp]
    \centering
    \includegraphics[height=10cm, width=18.5cm]{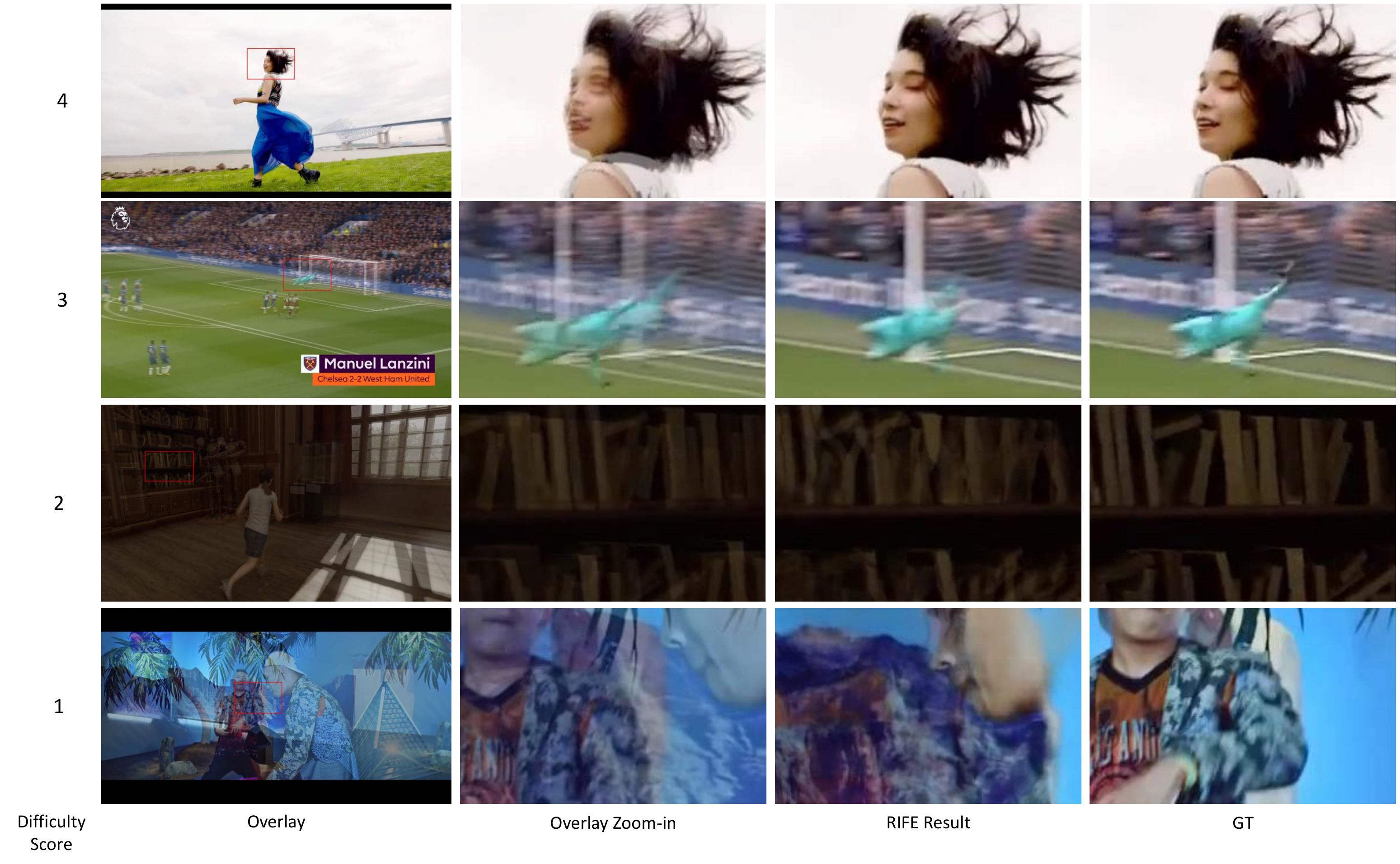}
    \caption{Example frames from our VFI Difficulty Assessment dataset. Top to down rows show easy to hard cases: higher difficulty score represents lower interpolation difficulty level.}
\label{fig:database}
\end{figure*}

\textbf{Auxiliary loss.} 
We argue that perceptual similarity of input frames has a positive correlation
with interpolation difficulty, because it can reflect motion magnitude and
dynamic texture complexity implicitly. Motivated by this, we introduce an
auxiliary loss to guide the difficulty prediction branch with perceptual similarity. Our auxiliary loss is defined as 
\begin{eqnarray}
    L_{AUX} = \|  S_{mean} - S_{lpips} \Vert 
\end{eqnarray}
where $S_{mean}$ is mean of spatial attention map from difficulty prediction branch in
Patch-wise prediction and $S_{lpips}$ is LPIPS~\cite{zhang2018unreasonable} of input images. We choose LPIPS as perceptual quality since it is
commonly used for accessing VFI methods.

\section{VFI difficulty assessment dataset}
Our VFI difficulty assessment dataset contains 13030 image triplets, where each triplet contains 3 neighboring frames and an interpolation difficulty score.

\textbf{Data preparation.}
We collect around 8000 video clips from Vimeo90K~\cite{xue2019video},
Adobe240~\cite{shen2020blurry} and Internet. These video clips contain three
resolutions (1080p, 720p, and 256p), and two frame rates (30 FPS
and 60 FPS). We extract image triplets from these video clips, and to stimulate large motion, we increase time interval
between triplets in some cases.

\textbf{Data annotation.} We assume the quality of interpolated result could
represent the interpolation difficulty of its input frames.
We interpolate each image triplet with RIFE~\cite{huang2020rife} and evaluate a perceptual quality score for RIFE's interpolation. The quality score ranges from 1 to 4 and used as interpolation difficulty level in our dataset. We provide illustration
of reference images from different difficulty levels in Fig.~\ref{fig:database}
It is shown that the difficult cases for frame interpolation typically
have large motions and rich texture.
In this work, we roughly categorize image pairs as easy and hard during inference, but it is worth noting that our dataset has 4 difficulty levels and supports more fine-grained model selection.

% \makeatletter
% \g@addto@macro\@maketitle{
%   \begin{figure}[H]
%   \setlength{\linewidth}{\textwidth}
%   \setlength{\hsize}{\textwidth}
%   \centering
%   \includegraphics[height=10cm, width=18.5cm]{dataset.pdf}
%   \caption{Example frames from our VFI Difficulty Assessment dataset. Top
%       to down rows show easy to hard cases: higher difficulty score represents
%   lower interpolation difficulty level.}
%   \label{fig:database}
%   \end{figure}
% }
% \makeatother

\begin{table*}[htbp]
    \begin{center}
    \caption{Dynamic VFI performance}\label{table:Dynamic VFI performance}
    \setlength{\tabcolsep}{11pt}
    \begin{tabular}{cccccc} \hline
    \multirow{2}{*}{Method} & \multicolumn{4}{c}{SUM-FLIM} &\multirow{2}{*}{Runtime} \\ \cline{2-5}
    { } & Easy & Medium & Hard & Extreme \\ \hline
    RIFE        & 40.0068/0.9904 & 35.7027/0.9785 & 30.0675/0.9327 & 24.8157/0.8530 & 0.016s \\ \hline
    Ours(256p)  & 40.0113/0.9904 & 35.7330/0.9786 & 30.3269/0.9352 & 25.2734/0.8622 & 0.355s \\ 
    Ours(720p)  & 40.0118/0.9904 & 35.7457/0.9787 & 30.3721/0.9355 & 25.3382/0.8630 & 0.443s \\
    Ours(1080p) & 40.0041/0.9904 & 35.7399/0.9787 & 30.2817/0.9349 & 25.2503/0.8623 & 0.348s \\ \hline
    VFIformer   & 40.1262/0.9907 & 36.0920/0.9798 & 30.6704/0.9378 & 25.4334/0.8642 & 1.268s \\ \hline
    \end{tabular}
    \end{center}
\end{table*}
\vspace{0.1pt}

\begin{table*}[t]
    \begin{center}
    \caption{VFI Per-Assessment performance}
    \label{table:VFI-PA-performance}
    \setlength{\tabcolsep}{6pt}
    \begin{tabular}{c|cccc|cc|cc} \hline
    {Method} & {Piexlshuffle} & \makecell{Lite-weight \\ Backbone} & \makecell{Image \\ Difference} & \makecell{Auxiliary \\ Loss} & {Tol.=0.125} & {Tol.=0.25} & {Para.} & {Runtime} \\\hline
    AHIQ~\cite{lao2022attentions} &{}            & {}         & {}     &            & 0.5507 & 0.8080 & 11.22M & 0.022s \\
    Baseline &{}       & {$\surd$}  & {}         &            & 0.4843 & 0.7305 & 4.490M & 0.014s \\
    {}   &{}           & {$\surd$}  & {$\surd$}  &            & 0.5117 & 0.8164 & 4.735M & 0.015s \\
    {}   &{$\surd$}    & {$\surd$}  & {}         &            & 0.5039 & 0.7500 & 4.233M & 0.014s \\
    {}   &{$\surd$}    & {$\surd$}  & {$\surd$}  &            & 0.5468 & 0.7539 & 4.469M & 0.014s \\
    Ours &{$\surd$}    & {$\surd$}  & {$\surd$}  & {$\surd$ } & 0.5546 & 0.7821 & 4.469M & 0.014s \\ \hline
    \end{tabular}
    \end{center}
\end{table*}

\section{Experiments}
\subsection{Experimental settings}
\textbf{Dataset.} We use SUN-FILM~\cite{choi2020channel} database to
evaluate dynamic VFI pipeline, and use VFI difficulty pre-assessment dataset to evaluate VFI-DPA. SUN-FILM is a public VFI dataset with 4 subsets, Easy, Medium, Hard and Extreme. These subsets are divided by motion magnitude and the average motion magnitude increases from Easy to Extreme.

\textbf{Evaluation Metrics.} We evaluate VFI-DPA on SUM-FILM by the accuracy under error tolerance.  Specifically, we calculate the absolute error between prediction and ground truth. If the error is less than an error tolerance threshold, the prediction is considered as True
Positive. In this work, we set two error tolerance thresholds: 0.125 and 0.25.
We evaluate dynamic VFI and compare it with static VFI using only RIFE or VFIformer. VFI performance is measured by peak signal to noise
ratio (PSNR) and structural similarity (SSIM).

\subsection{Main quantitative results.}
\textbf{Dynamic VFI performance.} 
As shown in Table~\ref{table:Dynamic VFI performance}, our pipeline surpasses RIFE by a large margin with small extra computational cost. 
We also observe that: 1) VFI-DPA model trained with different resolution datasets (256p, 720p, and 1080p) result in differences in accuracy-efficiency balance, and VFI-DPA trained with 720p dataset (same resolution with SNU-FILM) optimally improves the accuracy. 2) VFI-DPA better improves the accuracy on Hard and Extreme subsets, which is consistent with our motivation.

\textbf{VFI-DPA performance.}
As reported in table~\ref{table:VFI-PA-performance}, VFI-DPA achieves a better trade-off between accuracy and
efficiency than AHIQ. VFI-DPA maintains the accuracy of AHIQ with less
than half parameters. Its design can capture motion magnitude and pre-assess VFI difficulty successfully.

Our VFI-DPA matches the accuracy of AHIQ with less than half parameters, because its lightweight architecture is capable to capture the motion information that favors interpolation difficulty pre-assessment.

\subsection{Ablation study}
\textbf{Lightweight Backbone.} From Table~\ref{table:VFI-PA-performance}, 
our Siamese backbone reduces about 60\% of parameters with only 0.07 accuracy degradation, compared to original mixed CNN and ViT backbone adopted by AHIQ. Furthermore, we employ several tricks to match its accuracy with AHIQ.

\textbf{Effect of Image difference.} Image difference is a necessary information
for difficulty pre-assessment model. In Table~\ref{table:VFI-PA-performance},
it brings 0.03 improvement over baseline, which suggests VFI-DPA may
learn differences between input images at semantic level.

\textbf{Effect of Auxiliary Loss.} Our auxiliary loss for VFI-DPA is the MSE loss between LPIPS
similarity and mean of spatial attention map. As
reported in Table~\ref{table:VFI-PA-performance}, our auxiliary loss improves accuracy, demonstrating the positive
relationship between input images similarity and interpolation difficulty.

\textbf{Influence of Pixelshuffle layer.} We investigate the impact of
Pixelshuffle layer in DeformNet. We replace Pixelshuffle with traditional
interpolation. The comparison results in Table~\ref{table:VFI-PA-performance}
show that Pixelshuffle improves performance by 0.02. The Pixelshuffle can 
reconstruct more detailed spatial features than nearest
interpolation when rescaling feature from low resolution to high resolution.
We believe that Pixelshuffle layer has potential in image quality
assessment, where high resolution plays a key role.

\section{Conclusion} 
This paper presents a dynamic VFI pipeline, dedicated to
effectively generate pleasant intermediate frames with dynamic VFI model selection. The
results on SUM-FILM show that dynamic VFI achieves competitive performance with
less computational cost.
We believe that our dynamic VFI framework can provide an effective solution for many industrial applications which are concerned with accuracy-efficiency trade-off. Furthermore, our VFI-DPA model is a lightweight plug-and-play module which can enjoy the advances of VFI models to achieve better performance. Besides, we contribute a large scale Video Frame Interpolation Difficulty Assessment
dataset to train and validate VFI-DPA.

\end{document}